\documentclass[letterpaper, 10 pt, conference]{ieeeconf}  
\usepackage{amsmath,amsfonts}
\usepackage{algorithmic}
\usepackage{algorithm}
\usepackage{array}
\usepackage[caption=false,font=normalsize,labelfont=sf,textfont=sf]{subfig}
\usepackage{textcomp}
\usepackage{stfloats}
\usepackage{url}
\usepackage{verbatim}
\usepackage{graphicx}
\usepackage{cite}
\usepackage{xcolor}
\usepackage[left=1.91cm, right = 1.91cm, top=1.91cm, bottom=1.91cm]{geometry}
\let\labelindent\relax
\usepackage{enumitem}
\usepackage{hyperref}
\usepackage{multirow}
\usepackage{float}
\usepackage[font=normalsize,labelfont=bf]{caption}
\def\BibTeX{{\rm B\kern-.05em{\sc i\kern-.025em b}\kern-.08em
    T\kern-.1667em\lower.7ex\hbox{E}\kern-.125emX}}

\IEEEoverridecommandlockouts                              

\title{\LARGE \bf
A Chain-Driven, Sandwich-Legged Quadruped Robot: Design and Experimental Analysis
}

\author{Aman Singh$^{1}$, Bhavya Giri Goswami$^{2}$, Ketan Nehete$^{3}$, and Shishir N. Y. Kolathaya$^{1}$
\thanks{This work is supported by the SERB core research grant: CRG/2021/008115, and Pratiksha Trust Young Investigator Fellowship.}
\thanks{$^{1}$Aman Singh and Shishir N. Y. Kolathaya are with Department of Cyber Physical Systems, Indian Institute of Science, Bengaluru. {\tt\scriptsize \{saman, shishirk\}@iisc.ac.in}}
\thanks{$^{2}$ Bhavya Giri Goswami is with University of Waterloo, Waterloo, Ontario, Canada {\tt\scriptsize \{bhavyagg27@gmail.com\}}}
\thanks{$^{3}$ Ketan Nehete is with UC Berkeley, Berkeley, California, United States
{\tt\scriptsize \{ketan.nehete@berkeley.edu\}}}
}

\begin{document}

\newcommand{\circled}[1]{\textcircled{\scriptsize #1}}
\newcommand{\circledSmall}[1]{\raisebox{0pt}{\textcircled{\raisebox{0.6pt}{\tiny #1}}}}
\newcommand{\circledSuperSmall}[1]{\raisebox{0pt}{\textcircled{\raisebox{0.6pt}{\fontsize{4.2pt}{4.2pt}\selectfont #1}}}}

\maketitle
\thispagestyle{empty}
\pagestyle{empty}

\begin{abstract}
This paper introduces a chain-driven, sandwich-legged mid-size quadruped robot designed as an accessible research platform. The design prioritizes enhanced locomotion, improved actuation reliability and safety, and simplified, cost-effective manufacturing. Locomotion performance is improved through a sandwiched leg architecture and dual-motor configuration, reducing leg inertia for agile motion. Reliability and safety are enhanced using robust cable strain reliefs, motor heat sinks for thermal management, and mechanical limits to restrict leg motion. The design incorporates quasi-direct-drive (QDD) actuators and low-cost fabrication methods such as laser cutting and 3D printing for rapid prototyping. The $25\,\mathrm{kg}$ robot is built under \$8000, providing an affordable quadruped research platform. Experiments demonstrate trot and crawl gaits on flat terrain and slopes. We also open-source the mechanical designs.

VIDEO: \href{https://youtu.be/ygSMCPcFnP8?feature=shared}{Video}; 
CADs: \href{https://github.com/singhaman1750/stoch3-design.git}{Github}

\end{abstract}


\section{Introduction}

Legged robots have the potential to revolutionize real-world robotic applications with their ability to navigate unstructured terrains. This makes them invaluable for tasks such as power-plant inspections, transporting goods over uneven ground, and disaster response. Advancing reliable legged systems requires robust control algorithms and physical testing. Although quadrupedal platforms exist, further improvements in hardware design and accessibility are crucial to accelerate research and bridge theoretical models with practical deployment.

\begin{figure}[htbp]
    \centering
    \includegraphics[width=0.8\linewidth]{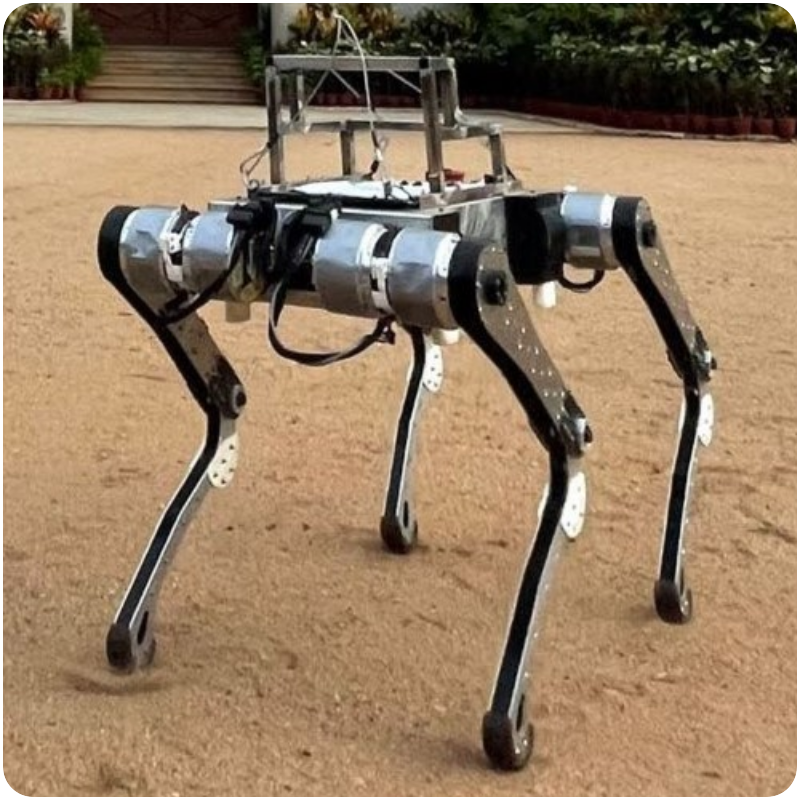} 
    \caption{Chain-driven, Sandwich-legged quadruped robot shown standing on the ground.}
    \label{fig:Stoch3}
\end{figure}

The MIT Cheetah-3\cite{MitCheetah3} demonstrates dynamic motions using high-torque-density custom motors, though requiring advanced manufacturing and high development costs. Similarly, the MIT Mini-Cheetah\cite{MiniCheetah} features a lightweight design with modular actuators integrating planetary gearboxes inside outrunner BLDC motors, enabling dynamic movements like backflips but requiring complex manufacturing processes.
Stanford Doggo \cite{StanfordDoggo} is an open-source, cost-effective quadruped with torque-controlled joints but limited by 2-DOF legs. Similarly, Mini-Taur \cite{MiniTaur} faces joule heating and restricted locomotion due to its gearbox-free, 2-DOF design.

The KAIST Hound \cite{KaistHound} uses parallel actuators for hip and knee joints, demanding precision manufacturing and raising costs. Panther \cite{PantherRFMPC}\cite{PantherLeg} employs simple compound planetary gearbox actuators but suffers from wiring complexity due to separate motor drivers.

\begin{figure}[htbp]
    \centering
    \includegraphics[width=0.95\linewidth]{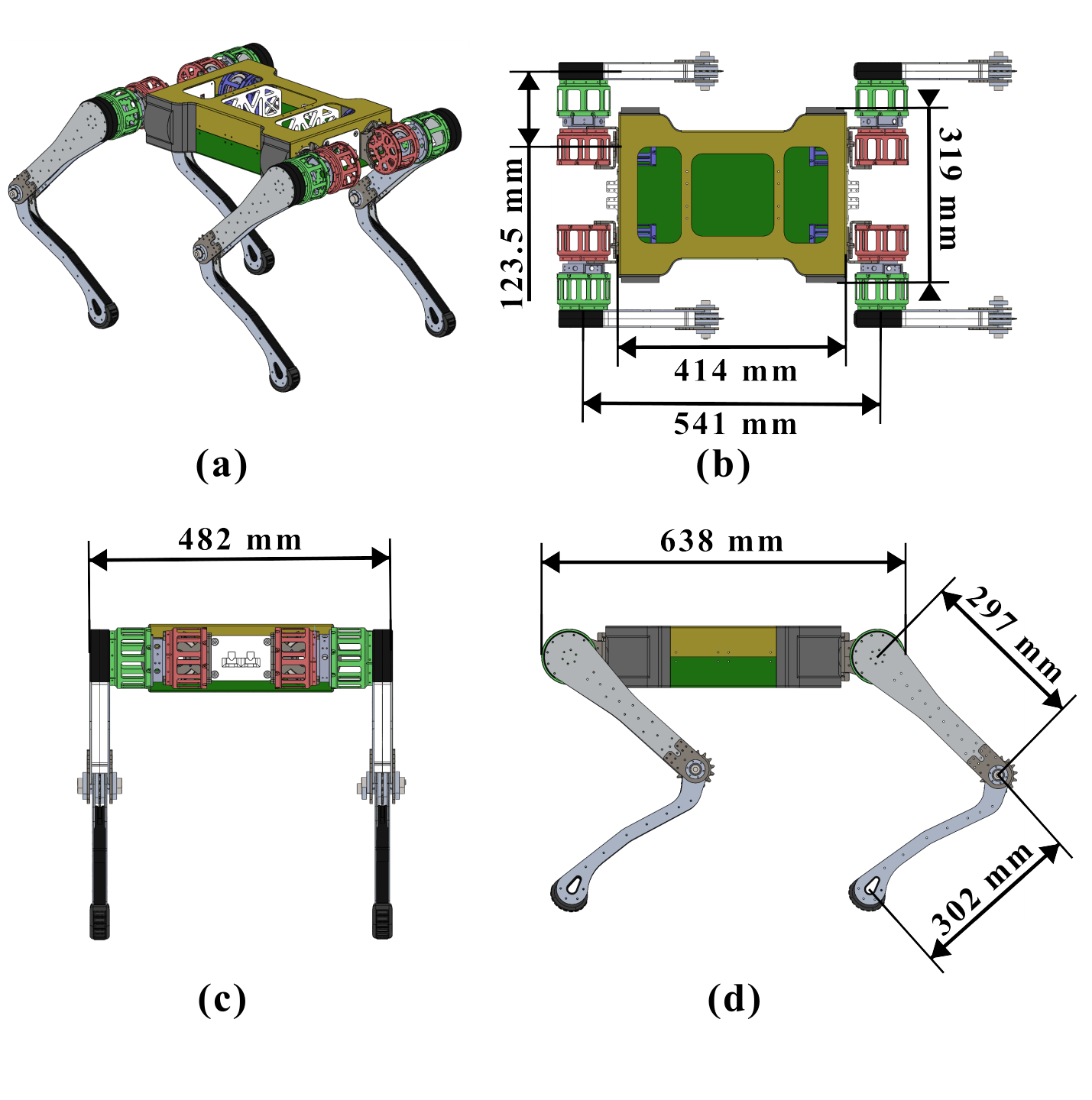} 
    \caption{\textbf{Robot Dimensions:} (a) Isometric view of the robot: The torso is made up of two sheet metal parts (yellow and green), with two plastic parts (white and inside the torso) joinging them (b) Top view, (c) Front View,  (d) Side view }
    \label{fig:RobotDimension}
\end{figure}

\begin{figure}[htbp]
    \centering
    \includegraphics[width=0.9\linewidth]{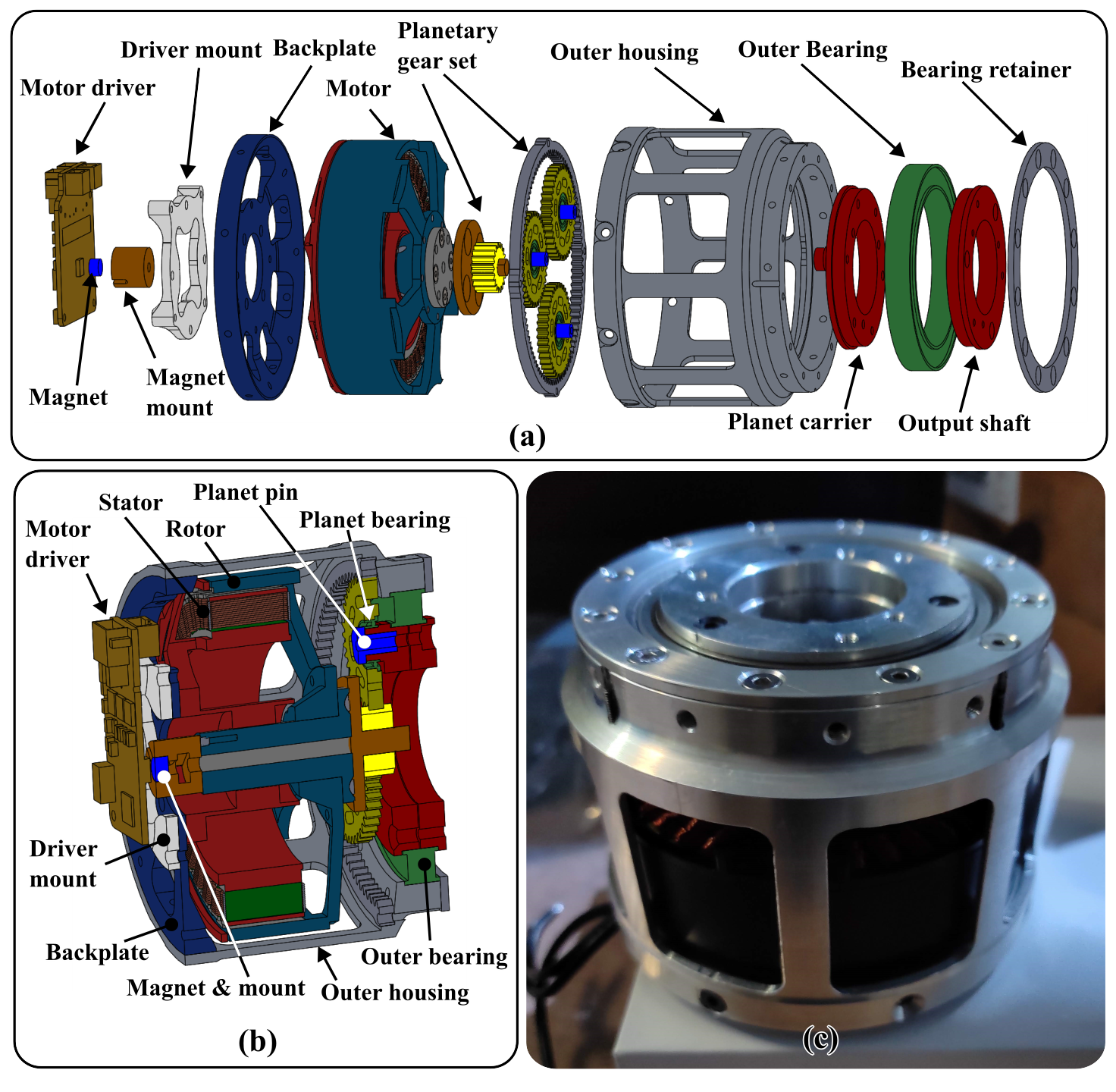} 
    \caption{\textbf{QDD Actuator:} (a) Details of the actuator design; (b) Cross-section view of actuator; (c) Manufactured actuator}
    \label{fig:Actuator}
\end{figure}

Hydraulic systems like HyQ \cite{HyQ} offer high power and force density for heavy-load tasks but are heavy and complex. Similarly, the ANYmal\cite{ANYmal} robot, employing harmonic drives and series elastic actuators, ensures precise torque control and impact resistance but sacrifices control bandwidth and simplicity, increasing manufacturing costs.
Open-source robots like Solo \cite{Solo} use simplified 3D-printed and off-the-shelf components, enhancing accessibility but suffering from belt slippage under high loads. Barry \cite{Barry} supports high payloads but uses complex ball-screw transmissions, compromising design simplicity.
Finally, commercial platforms like Spot\cite{Spot}, Vision60\cite{Vision60}, and Go1\cite{go1} offer reliability but lack publicly available design details. Additionally, Spot and Vision60 are prohibitively expensive, limiting their adoption for research purposes.

From the analysis, as shown in Table \ref{table:quadrobot_perf}, many quadrupedal platforms demonstrate impressive locomotion but involve trade-offs. Systems like \cite{MitCheetah3}, \cite{MiniCheetah}, \cite{KaistHound}, \cite{Barry}, and \cite{ANYmal} require costly precision manufacturing. Simpler robots like \cite{StanfordDoggo} and \cite{MiniTaur} have limited 2-DOF legs, while \cite{Solo} and \cite{PantherRFMPC} face reliability challenges. Commercial robots often remain expensive or lack open designs.

This paper presents a quasi-direct drive quadruped with a chain-sprocket knee drive, detailing its hardware development, locomotion-enhancing design choices, actuator reliability improvements, and low-cost, rapid manufacturing principles.

The main contributions of the paper are as follows.
\begin{itemize}
    \item A chain-driven quasi-direct drive actuator with an integrated planetary gearbox and motor driver, eliminating the need for machining off-the-shelf or frameless motors \cite{MiniCheetah, Barry, KaistHound}.

    \item A sandwiched leg design using a lightweight composite core between laser-cut metal sheets, reducing inertia and manufacturing cost, offering an alternative to CNC-intensive methods \cite{MitCheetah3, PantherLeg, ANYmal, MiniCheetah, HyQ, KaistHound, Barry}.

    \item A cost-effective, strong torso structure using sheet metal and plastic.
\end{itemize}

The paper is organized as follows: Section \ref{sec:Design} covers the robot’s design and principles; Section \ref{sec:ElectronicsAndControl} details electronics and control; Section \ref{sec:Results} presents hardware results; Section \ref{sec:Conclusion} concludes and outlines future work.

\nocite{}

\begin{table*}[!t]
\caption{Overview of Quadruped Robots’ Design}
\label{table:quadrobot_perf}
\centering
\begin{tabular}{l c c c c c c c}
\hline
\textbf{Name} & \textbf{Type} & \textbf{Leg Size [m]} & \textbf{Mass [kg]} & \textbf{Open Source Design} & \textbf{Leg DOF} & \textbf{Design Remarks} & \textbf{Cost} \\
\hline
SOLO \cite{Solo}           & Elec. & 0.32 & 2.2 & Yes & 3 & Belt slippage issue & \$ \\
Stanford-Doggo \cite{StanfordDoggo} & Elec. & 0.32 & 4.8 & Yes & 2 & Simplified Design & \$ \\
Minitaur \cite{MiniTaur}       & Elec. & 0.4 & 5.0 & Yes & 2 & Simplified Design & \$ \\
Panther \cite{PantherRFMPC}\cite{PantherLeg}        & Elec. & 0.28 & 5.5 & Yes & 3 & Separate driver and actuator & \$ \\
Mini-Cheetah \cite{MiniCheetah}   & Elec. & 0.4 & 9.0 & Yes & 3 & Complex Gearbox design  & \$ \\
MIT cheetah 3 \cite{MitCheetah3}  & Elec. & 0.68 & 45.0 & Yes & 3 & Complex Motor Design  & \$\$ \\
KAIST Hound \cite{KaistHound}    & Elec. & 0.68 & 45.0 & Yes & 3 & Parallel config. in Hip \& Knee & \$\$ \\
HyQ \cite{HyQ}            & Hydr. & 0.7 & 91 & Yes   & 3 & Hydraulic based actuation & \$\$ \\
Barry \cite{Barry}          & Elec. & 0.8 & 48.0 & Yes & 3 & Complex Screw knee trans. & \$\$ \\
Unitree Go-1 \cite{go1}   & Elec. & 0.42 & 12 & No  & 3 & - & \$ \\
ANYmal \cite{ANYmal}         & Elec. & 0.5 & 30.0 & No  & 3 & Direct drive knee transmission & \$\$\$ \\
Spot \cite{Spot}           & Elec. & 0.75 & 32 & No  & 3 & - & \$\$\$ \\
Vision-60 \cite{Vision60}      & Elec. & 0.68 & 51 & No  & 3 & - & \$\$\$ \\
Stoch-3 (This Paper)   & Elec. & 0.6  & 25 & Yes & 3 & Simplified design & \$ \\
\hline
\end{tabular}
\footnotetext[1]{Estimated from other parameters in this table or in the work cited.}
\footnotetext[2]{Maximum value, not used for the estimation of other parameters.}
\footnotetext[3]{Not including power consumption due to computation.}
\footnotetext[4]{Theoretical, from design specs or max joint torque, but not demonstrated in walking.}
\footnotetext[5]{Not used for estimation of other parameters.}
\footnotetext[6]{Autonomy estimate.}
\end{table*}

\begin{figure}[htbp]
    \centering
    \includegraphics[width=\linewidth]{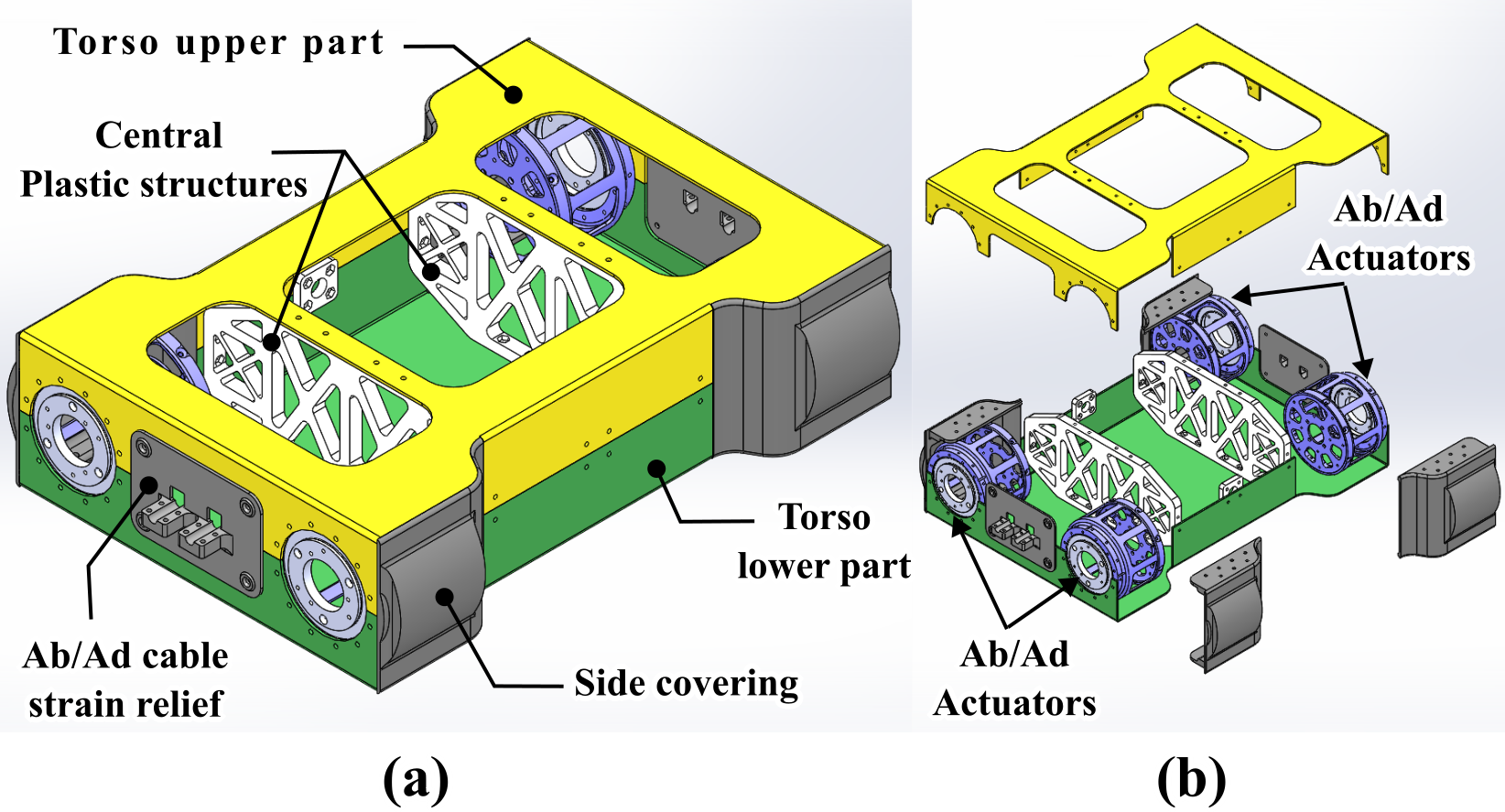} 
    \caption{\textbf{Torso Diagram:} (a) Details of Torso design (b) Exploded view of Torso }
    \label{fig:TorsoDiagram}
\end{figure}

\section{Design} \label{sec:Design}
This section describes the design of the custom quadruped robot shown in Fig.~\ref{fig:Stoch3}. Each leg has three actuated degrees of freedom, driven by a modular Quasi-Direct-Drive (QDD) actuator with a single-stage planetary gearbox. The robot lacks foot contact or force sensors and detects ground contact using joint actuator currents. As shown in \cite{MitCheetah3}, up to $99.4\%$ ground contact detection accuracy is achievable using joint actuator currents. The specific detection algorithm is detailed in \cite{Stoch3LinearPolicy}. Robot dimensions are shown in Fig.~\ref{fig:RobotDimension}, with a total weight of approximately 25~kg. Design decisions to enhance locomotion, actuator reliability, and manufacturing simplicity are discussed.

\subsection{Simplified Design}
\subsubsection{QDD Actuator}

The robot utilizes a custom-built quasi-direct-drive actuator, featuring an off-the-shelf BLDC motor (T-motor, U10 plus, Kv100) paired with a 6:1 single-stage planetary gear reduction and a moteus r4.10 motor driver from MJBOTS \cite{mjbots} (Fig. \ref{fig:Actuator}). The actuator incorporates KHK gears with a 0.8 mm module, where the sun gear has 20 teeth, the planet gear has 40 teeth, and the ring gear has 100 teeth. Post-purchase machining was performed to customize the gears. Weighing approximately 920 g, the actuator's motor driver includes an onboard Hall effect sensor, allowing it to be mounted directly on the actuator module for a compact design and reduced wiring.

\subsubsection{Design for efficient manufacturing}
To minimize manufacturing cost and time, the robot's components were designed to enable efficient fabrication processes without compromising structural integrity. The design extensively utilizes metal laser cutting and 3D printing for fast production.
The torso consists of a cuboid-shaped box made from two sheet metal parts, which are laser cut and bent (Fig. \ref{fig:TorsoDiagram}). These two sheet metal parts are connected by two 3D-printed rectangular components, and the abduction actuators are mounted at each corner to provide additional strength. This design reduces fabrication time and cost while maintaining structural strength. The 3D-printed parts inside the torso also serve as mounting points for the electronics.

\begin{figure*}[h]
    \centering
    \includegraphics[width=\textwidth]{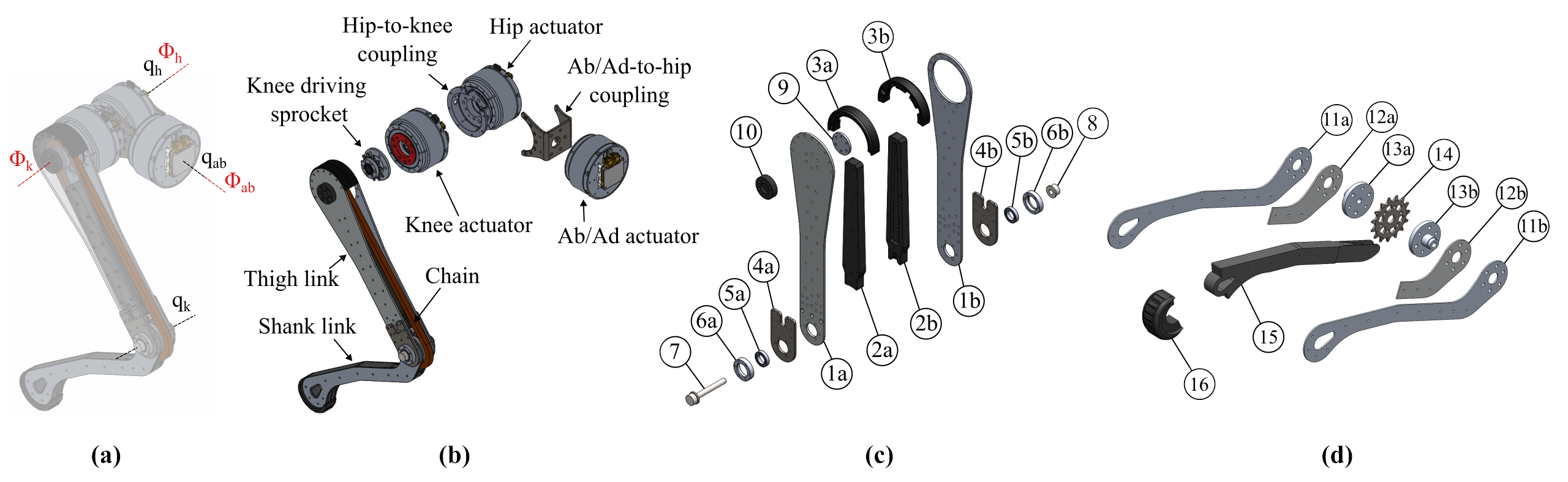}
    \caption{\textbf{(a) Joints and Actuator Axes; (b) Details of Leg Design; (c) Details of Thigh Link:} \circled{1a}, \circled{1b}: Al Sheets; \circled{2a}, \circled{2b}: Central spacers; \circled{3a}, \circled{3b}: Upper spacers; \circled{4a}, \circled{4b}: Thigh strength plates; \circled{5a}, \circled{5b}: Knee bearings; \circled{6a}, \circled{6b}: Knee bearing blocks; \circled{7}: Knee bolt; \circled{8}: Knee nut; \circled{9}: Sprocket support shaft; \circled{10}: Support shaft fastener\textbf{; (d) Details of Shank Link:} \circledSmall{11a}, \circledSmall{11b}: Shank Al-sheets; \circledSmall{12a}, \circledSmall{12b}: Shank strength plates; \circledSmall{13a}, \circledSmall{13b}: Knee sprocket couplings; \circled{14}: Knee driven sprocket; \circled{15}: Shank spacer; \circled{16}: Foot.}
    \label{fig:LegDesign}
\end{figure*}

\begin{figure}[htbp]
    \centering
    \includegraphics[width=0.8\linewidth]{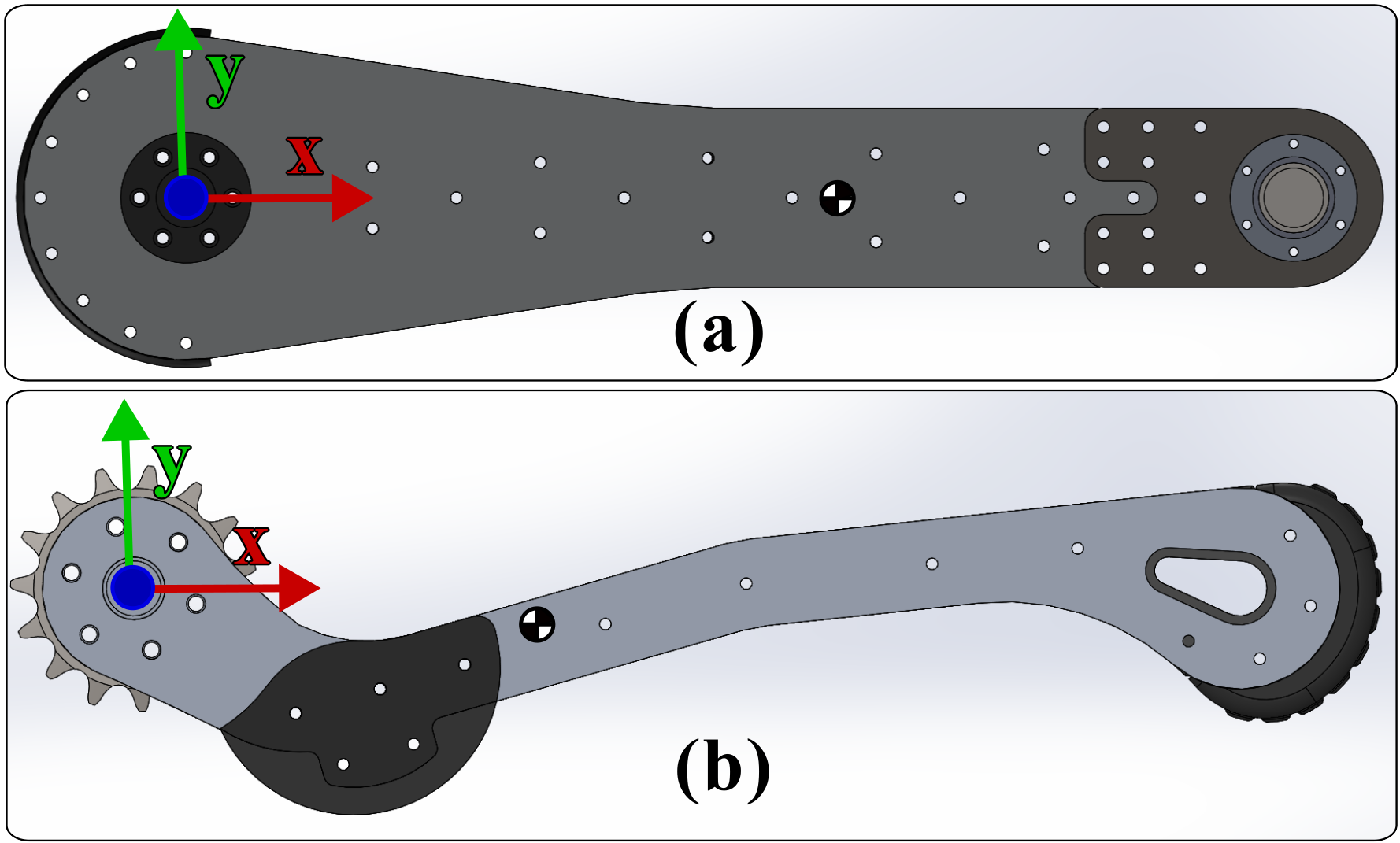} 
    \caption{CoM locations and coordinate systems used for calculating the moments of inertia (MoI) of the thigh and shank links.}
    \label{fig:Link_COM_MOI_diagram}
\end{figure}

As discussed in the leg design section, the robot’s legs use laser-cut metal for strength and 3D-printed plastic for spacing and shape definition (Fig.~\ref{fig:LegDesign}). The abduction-to-hip coupling, critical for load-bearing, and the knee chain tensioner are both fabricated from laser-cut steel, ensuring durability and enabling rapid manufacturing.

\subsection{Enhancement of Locomotion Capability}

Low leg inertia enables faster leg movements and reduces actuator effort during walking. To achieve this, we adopt a sandwich leg design and dual motor architecture, detailed in the following sections.

\begin{figure}[htbp]
    \centering
    \includegraphics[width=0.9\linewidth]{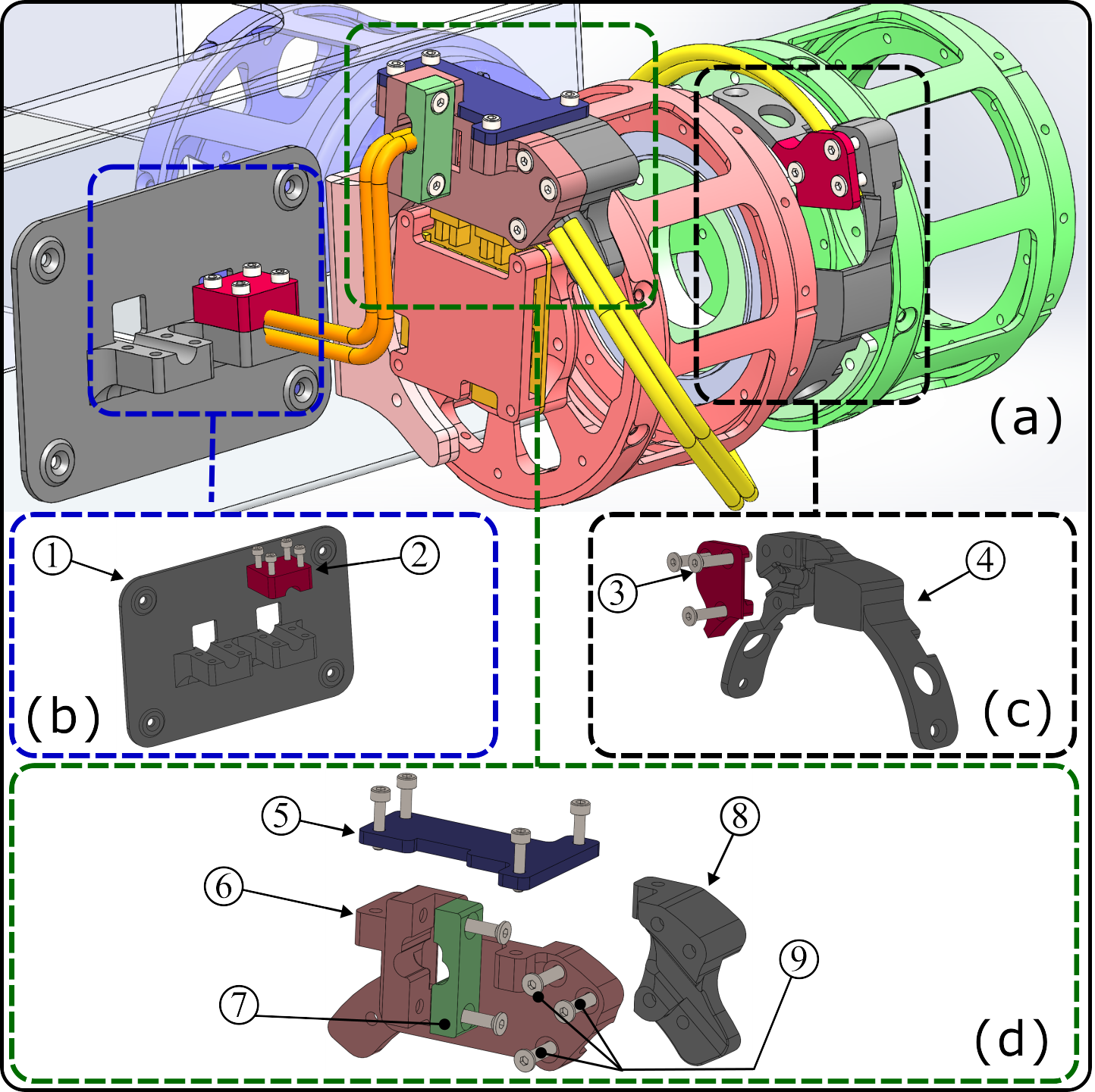} 
    \caption{\textbf{Cable Strain Reliefs:} (a) Anatomy of cables and strain reliefs; (b) Ab/Ad strain relief: \circled{1} mounted on the torso, \circled{2} fastens the Ab-Ad actuator wire (orange); (c) Knee strain relief: \circled{4} mounted on the hip-to-knee coupling, \circled{3} fastens the Hip actuator wire (yellow); (d) Hip strain relief: \circled{8} and \circled{9} mounted on the hip actuator, with fasteners \circled{6} and \circled{9} for the Knee actuator wire (yellow), \circled{7} for the Ab-Ad actuator wire (orange), and \circled{5} supporting from the top.}

    \label{fig:StrainReliefs}
\end{figure}

\begin{table}[ht]
\centering
\caption{Mass, CoM, and MoI of leg links. MoI is computed about the rotation center shown in Fig.~\ref{fig:Link_COM_MOI_diagram}.}
\begin{tabular}{c c c c}
\hline
\textbf{Link} & \textbf{Parameter} & \textbf{Value} & \textbf{Units} \\
\hline

\textbf{Thigh} & Mass & 0.6353 & kg \\
\textbf{Link}  & CoM & (0.1760, 0.0000, 0.0011) & m \\
(without       & $(I_{xx},I_{yy},I_{zz})$ & (0.000324, 0.028041, 0.028040) & kg$\cdot$m$^2$ \\
chain)       & $(I_{xy},I_{xz},I_{yz})$ & (0.0, -0.000009, 0.0) & kg$\cdot$m$^2$ \\
 \hline
 
\textbf{Shank} & Mass & 0.6109 & kg \\
\textbf{Link}  & CoM & (0.1088, -0.0096, 0.0000) & m \\
      & $(I_{xx},I_{yy},I_{zz})$ & (0.000326, 0.015362, 0.015607) & kg$\cdot$m$^2$ \\
      & $(I_{xy},I_{xz},I_{yz})$ & (-0.000328, 0.0, 0.0) & kg$\cdot$m$^2$ \\
\hline

\hline
\end{tabular}
\label{tab:compact_leg_properties}
\end{table}

\subsubsection{Sandwiched Leg Design}
The robot's leg consists of three links: the abduction link, the thigh link, and the shank link, as shown in Fig. \ref{fig:LegDesign}(b). The abduction link includes the abduction/adduction (Ab/Ad)-to-hip-coupling and the outer shell of the hip actuator. The thigh and shank links are designed using a "sandwich" structure of sheet metal and plastic.
The sandwich leg design incorporates two sheet metal plates with a 3D-printed plastic component positioned between them, as illustrated in Fig. \ref{fig:LegDesign}(c,d). The metal plates provide structural strength, while the plastic part contributes to the form and shape of the legs and serves as a spacer. These components are securely joined using superglue and fasteners.
To address the high loads experienced at the knee joint during locomotion, the regions near the knee joint in the thigh and shank links are reinforced with steel plates, as shown in Fig. \ref{fig:LegDesign}(c,d). This lightweight design reduces leg inertia, enhancing locomotion performance, and aligns with the simplified design philosophy discussed earlier. The mass and inertia values of the leg links are given in Table \ref{tab:compact_leg_properties}.

\subsubsection{Dual Motor Design}
The dual motor architecture \cite{DesignPrinciples}, inspired by animals like horses, places both hip and knee actuators near the hip joint. This design eliminates the need for a knee-mounted actuator, significantly reducing leg inertia and improving locomotion efficiency.
The knee joint is powered through a chain mechanism, where the rotor of the hip actuator is coupled to the stator of the knee actuator mounted on the thigh link. The knee actuator drives a sprocket that transmits power via the chain, also providing additional gear reduction by varying sprocket sizes. With an 11-tooth driving sprocket and a 15-tooth driven sprocket, the combined planetary and chain system achieves an effective gear reduction of 8.18:1, enhancing the knee joint’s torque during locomotion.

The chain drive is used as a transmission instead of alternatives like belt drives~\cite{MiniCheetah}, direct-drive actuators~\cite{ANYmal}, four-bar linkages~\cite{go1}, and screw-based mechanisms~\cite{Spot}. Belt drives, though lightweight and simple, require friction-based tensioners, increasing friction and efficiency losses. In contrast, chain drives use tension-based tensioners, avoiding friction and improving efficiency. Direct mounting of actuators increases leg inertia and control effort. Four-bar linkages compactly transmit motion but limit knee range to less than $90^\circ$, while screw mechanisms add design complexity and restrict motion. Thus, chain drives offer a balanced trade-off in efficiency, simplicity, and mechanical performance.

\subsection{Reliability and safety of actuation system}

The robot was designed for indoor and outdoor research. On rough terrains, erratic motion may disconnect actuator wires, while extended outdoor locomotion can overheat motor drivers, degrading performance or causing failure. Additionally, testing new control algorithms poses risks to nearby researchers. To address these issues, three measures were implemented to improve actuator reliability and safety.

\subsubsection{Cable Strain Relief}
Cable strain reliefs secure actuator wires and prevent disconnections during motion. Each leg uses a daisy-chain wiring configuration: instead of separate wires to each actuator, wires are routed sequentially through the abduction, hip, and knee actuators from the main electronics.

\begin{figure}[htbp]
    \centering
    \includegraphics[width=\linewidth]{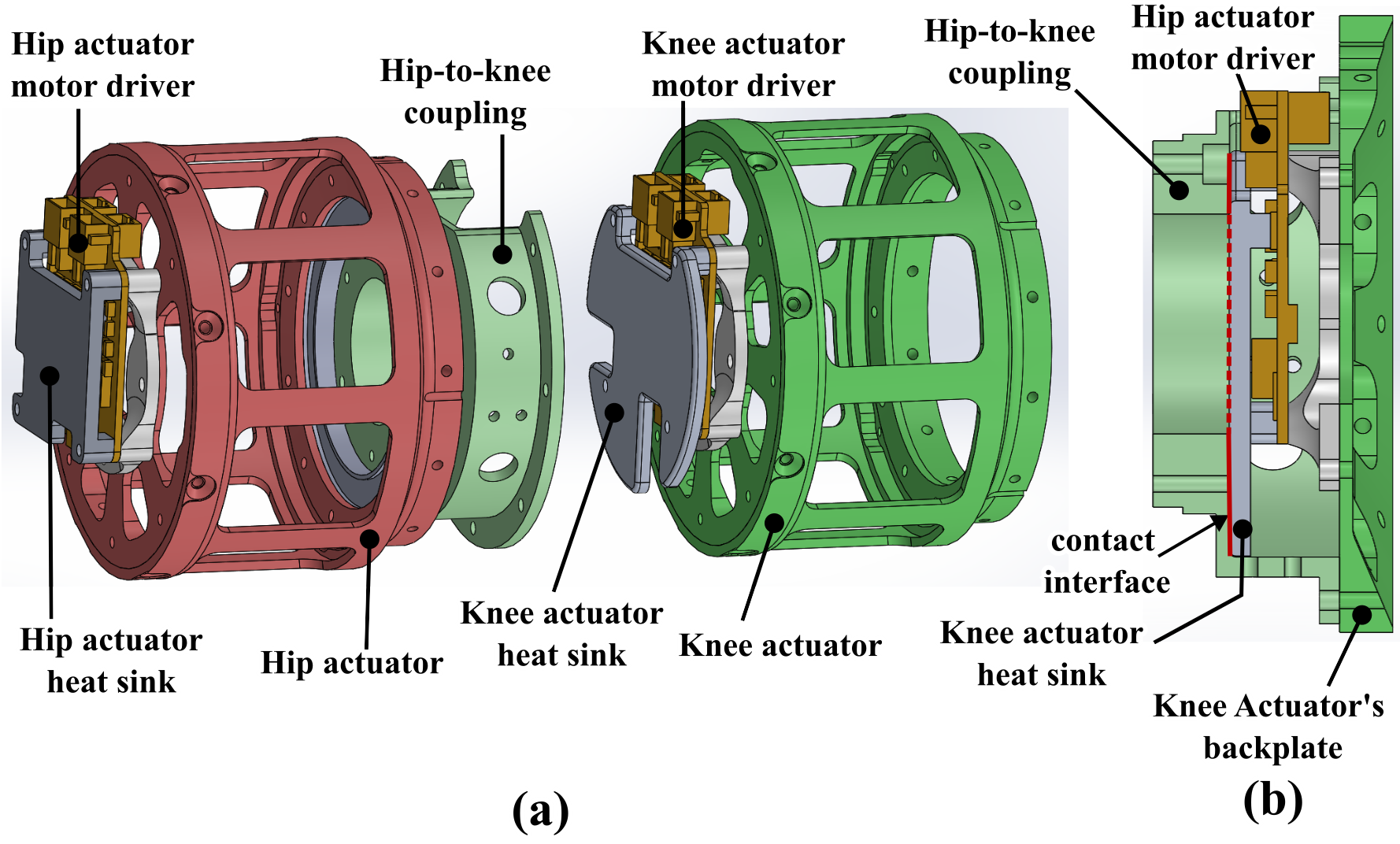} 
    \caption{\textbf{Thermal Management:} (a) Exploded view: The knee heat sink goes inside the cavity of the Hip-to-knee coupling; (b) Cross-section view: The knee heat sink touches the Hip-to-knee coupling at the contact interface (red).}
    \label{fig:ThermalManagement}
\end{figure}

The abduction actuator’s strain relief is mounted on the torso. Wires from the main electronics connect to the abduction actuator and pass through a hole in the strain relief, which is fastened to the torso (Fig.~\ref{fig:StrainReliefs}(a,b)). The hip strain relief secures two connections: one from the abduction actuator and another leading to the knee actuator (Fig.~\ref{fig:StrainReliefs}(a,d)). The knee strain relief, mounted on the hip-to-knee coupling, secures a single wire from the hip actuator. The knee motor driver is housed within the coupling cavity, with wires routed through a hole beneath the strain relief (Fig.~\ref{fig:StrainReliefs}(a,c); Fig.\ref{fig:ThermalManagement}(a)).

\subsubsection{Thermal Management}
Off-the-shelf heat sinks from \cite{mjbots} were used for the abduction and hip actuators. Custom-designed heat sinks were fabricated for the knee actuators, as their motor drivers overheated, reducing the robot’s uptime. The knee heat sink (Fig.~\ref{fig:ThermalManagement}) contacts the MOSFETs and driver board on one side and the hip-to-knee coupling on the other, which connects to the actuator’s outer shell, increasing thermal capacity. Thermal paste was applied at both interfaces to enhance conductivity. The heat sink also protects the previously exposed hip motor driver from physical damage.

\subsubsection{Safety Limits}
\begin{figure}[htbp]
    \centering
    \includegraphics[width=\linewidth]{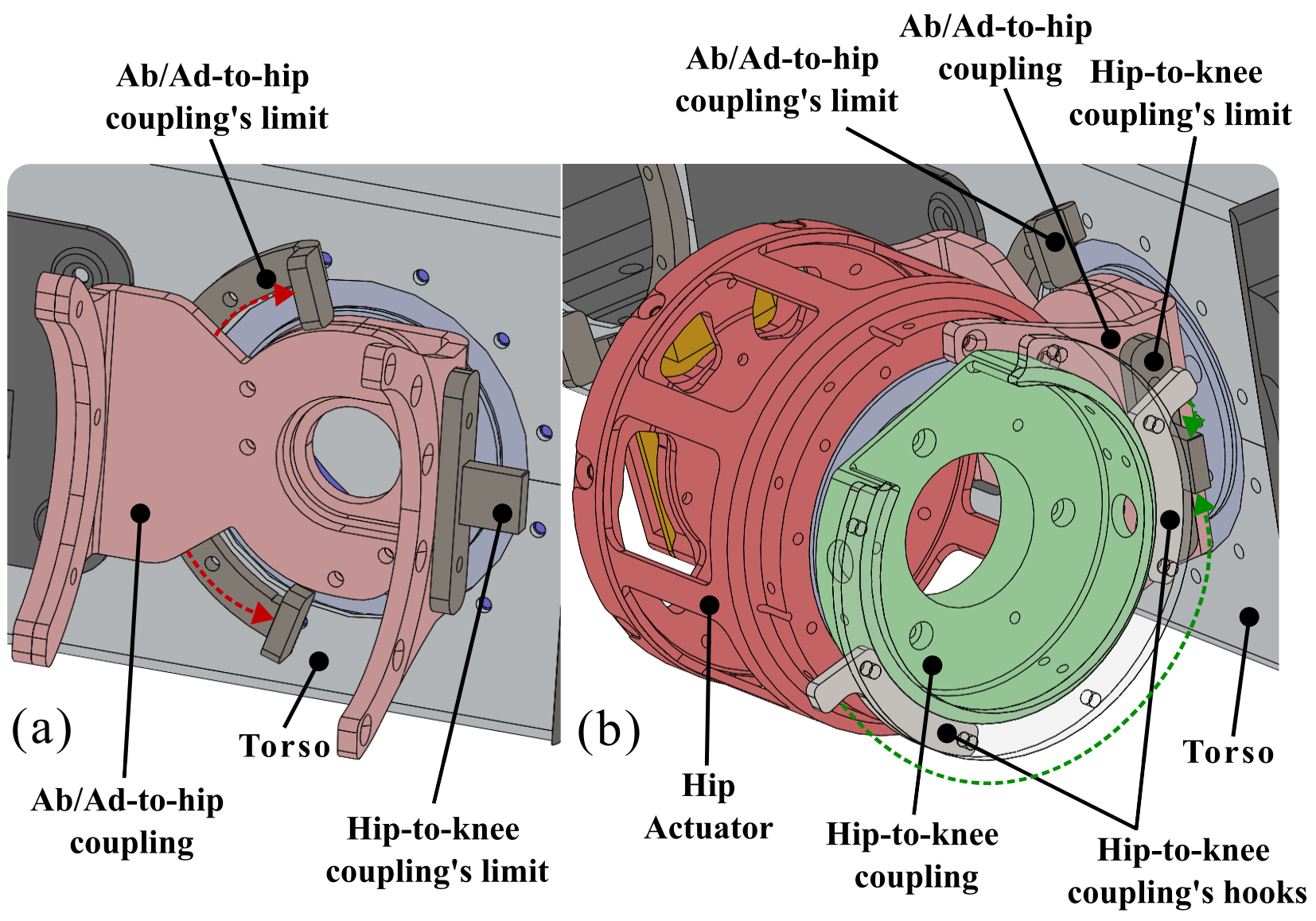} 
    \caption{\textbf{Safety Limits:} (a) Ab/Ad-to-hip coupling limit; (b) Hip-to-knee coupling limit}
    \label{fig:SafetyLimits}
\end{figure}

Mechanical safety limits are integrated into each leg joint to restrict motion and ensure operator safety during testing. The Ab/Ad joint is limited by the Ab/Ad-to-hip coupling limit, and the hip joint by the hip-to-knee coupling limit, as shown in Fig.~\ref{fig:SafetyLimits}. The Ab/Ad limit is fixed to the torso, while the hip limit attaches to the Ab/Ad-to-hip coupling. Hooks on the hip-to-knee coupling engage with the hip limit. All limits and hooks are made from laser-cut, bent steel sheets. The knee joint is constrained by the thigh link. Joint ranges are $\pm35^\circ$ (Ab/Ad), $+75^\circ$ to $-60^\circ$ (hip), and $+165^\circ$ to $-55^\circ$ (knee). These ranges are extendable through minor design modifications, offering flexibility for future improvements.
\section{Electronics and Control Architecture} \label{sec:ElectronicsAndControl}

Locomotion control is executed on a Raspberry Pi 4B (RPi) paired with an mjbots pi3hat board, handling communication with motor drivers. Four independent CAN buses manage communication with three actuators per leg. Each actuator uses an mjbots moteus r4.10 motor driver with onboard Hall-effect encoders for angle and velocity measurements via CAN. For orientation and motion sensing, the robot employs an Xsens MTi-610 Inertial Measurement Unit (IMU), providing calibrated 3D orientation, angular velocities, and accelerations.
The robot is powered by a $22.2$V, $10,000$~mAh Li-Po battery housed in a sliding box with a nut-bolt interface for easy swapping. Power distribution, managed by two off-the-shelf mjbots boards, uses a daisy-chain configuration. The torso also has space for an additional onboard computer for high-level tasks like image processing.
The control architecture, inspired by \cite{LinearPolicy}, uses a linear policy and lower-level controller. The linear policy generates trajectory modifications and body-wrench values, which are converted into joint torque commands. The setup evaluates actuator torque control. Further details are provided in \cite{Stoch3LinearPolicy}.

\section{Results} \label{sec:Results}
This section presents the results of the design decisions discussed earlier, hardware walking experiments, and potential design improvements identified through extensive testing of the robot.

\subsection{Design Principles Results}
This subsection highlights the effects of key design decisions on the reliability and performance of the robot.

\subsubsection{Simplified Design:}
The simplified design principles utilized sheet metal cutting and bending, combined with 3D-printed components, to construct the robot's torso and legs. Using off-the-shelf motors, gears, and drivers kept costs below \$8000, with approximately \$4500 for motors and \$3500 for electronics and manufactured parts, making it a cost-effective midsized robot platform.

\subsubsection{Enhancement of Locomotion Capability}
The sandwiched leg design and dual-motor architecture reduced leg inertia, enhancing locomotion capability. Each leg’s lightweight design ($1.5$ kg/leg) enabled faster control responses and effective movement across diverse terrains.

\subsubsection{Reliability and Safety of the Actuation System}
Key design improvements enhanced actuator reliability and safety. Cable strain reliefs resolved early disconnection issues during locomotion. Thermal coupling of the knee actuator to the hip-to-knee link increased runtime from 10 to 30 minutes. Mechanical joint limits ensured operator and robot safety during operation.

\subsection{Hardware Experiment Results}
\begin{figure}[htbp]
    \centering
    \includegraphics[width=\linewidth]{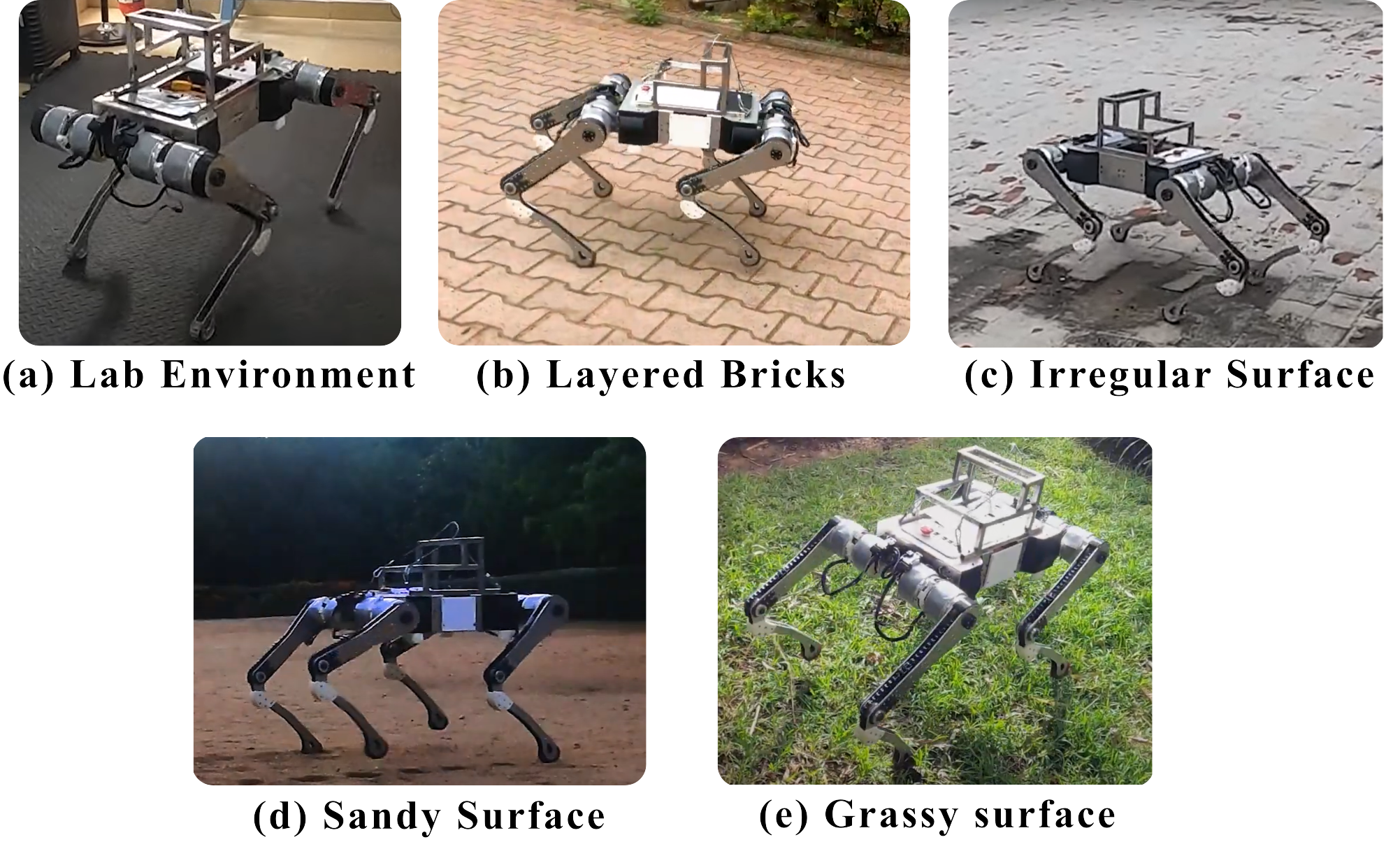}
    \caption{Robot walking on different surfaces}
    \label{fig:RobotWalking}
\end{figure}

The platform was tested in real-world scenarios, including laboratory experiments, outdoor locomotion on uneven terrains (sand and grass) with a trot gait, and handling external disturbances (Fig.~\ref{fig:RobotWalking}).  A video demonstration of the walking experiments is available at: \url{https://youtu.be/ygSMCPcFnP8?feature=shared}.

\section{Conclusion and Future Work} \label{sec:Conclusion}
This paper presented the design of a chain-driven, sandwich-legged quadruped robot, emphasizing simplicity, reliable actuation, and improved locomotion. Cost-effective manufacturing was achieved using QDD actuators, sheet metal, and 3D printing. Key features like dual-motor architecture and sandwiched legs enhanced performance, while reliability was ensured via strain reliefs, thermal management, and safety limits. Experiments validated outdoor terrain traversal and disturbance rejection.

Testing feedback highlighted improvement areas for future work. The current torso, made of two sheet metal parts with abduction actuators, bends under high forces at the mounting points. A thicker metal plate will enhance structural integrity. The knee actuator, which bears high loads, causes motor driver overheating. Thermal management can be improved by integrating the knee heat sink and hip-to-knee coupling into one component and adding fins for increased surface area. The current silicone-molded foot, though load-bearing, wears quickly on abrasive terrain. Future designs will use vulcanized rubber to improve durability and longevity in outdoor environments.

\bibliographystyle{IEEEtran} 
\bibliography{references} 

\end{document}